\documentclass[11pt]{article}

\usepackage[preprint]{acl}
\usepackage{times}
\usepackage{latexsym}
\usepackage[T1]{fontenc}
\usepackage[utf8]{inputenc}
\usepackage{microtype}
\usepackage{inconsolata}
\usepackage{graphicx}
\usepackage{booktabs}

\title{FriendBench: Benchmarking Dyadic Familiarity Inference\\in Humans and Multimodal Large Language Models}

\author{
  Jeffrey M. Girard, Jason Z. Zheng, Jacqueline R. Vertino,\\
  \textbf{Antony D'Avirro, Benjamin Peloquin} \\
  Fluid Concepts Research \\
  {\small \textbf{Correspondence:} \href{mailto:jeff@fluidconcepts.ai}{jeff@fluidconcepts.ai}}
}

\begin{document}
\maketitle
\begin{abstract}
Reading a social situation often depends on behavior, not words alone. We introduce FriendBench, a benchmark for inferring whether two people are already familiar or are meeting as strangers, from a 20-second clip of a dyadic ice-breaker conversation. Every pair answers the same type of prompt, so only the manner of interaction can reveal the answer. Across text, audio, and video, we compare 26 models from seven companies against matched human panels over 96 balanced dyads. The best model and the human crowd are statistically indistinguishable on accuracy in every modality, but reach it differently: humans stay balanced across the two answers, while the strongest models favor ``stranger.'' This is a difference in effective prior, not in discrimination. Richer channels help both unequally, and only humans gain from visible behavior on top of speech. We release the stimuli, human ratings, and model predictions.
\end{abstract}

\graphicspath{{figures/}{docs/figures/}}

\newcommand{\mlogo}[1]{\makebox[2.9ex][l]{\raisebox{-0.28\height}{%
  \includegraphics[height=1.7ex,width=2.6ex,keepaspectratio]{#1}}}}

\section{Introduction}
\label{sec:intro}
Social intelligence, the capacity to make sense of other people and the situations they are in, is a central component of human cognition \citep{adolphs2003,frith2007}. It is also an increasingly important requirement for AI systems, which now observe, mediate, and participate in human interaction in roles ranging from companion agents to meeting assistants and care monitors \citep{mathur2024,sap2019}. Such systems must reason about the social situation people are in, and much of that reasoning rests on multimodal behavioral cues rather than explicit verbal content, such as how people coordinate, respond, and orient toward one another \citep{tickle-degnen1990}. A system that reads only the semantic content of a conversation captures just part of what social understanding requires.

We study one concrete instance of social intelligence: the capacity to recognize the relationship between two people from a brief sample of how they interact. Specifically, we ask whether an observer can tell that two people are already familiar with one another rather than meeting as strangers, without being told and without relying on what the conversation is about. This judgment is a clean probe of behavioral social perception for two reasons. First, since we have ground-truth relationship labels, its answer is a matter of fact rather than interpretation: much related work in social intelligence targets intentions, emotions, or beliefs \citep{premack1978,sap2019}, whose correct answer is often contestable, whereas prior familiarity has an unambiguous ground truth (i.e., two people either have met before or they have not). Second, familiarity is often unstated in a short interaction, so it must be inferred from behavior. The task therefore rules out succeeding by reading semantic content alone, and prior work shows that humans make accurate social judgments from brief samples of behavior, so-called \emph{thin slices} \citep{ambady1992}.

We operationalize the task as \textsc{FriendBench}: binary classification, familiar versus strangers, from a 20-second clip of a dyadic ice-breaker conversation, evaluated separately across text, audio, and video modalities. Every dyad, whether familiar or strangers, responds to the same type of ice-breaker prompt, so the conversational topic is matched across the two classes and cannot by itself reveal the answer, leaving the \textit{manner} in which the two people interact as the primary signal.

In addition to the core capability we benchmark, we examine how humans and models draw on the three modalities, and how each reaches its accuracy: by telling the two classes apart (\emph{discrimination}) or by answering one class more often regardless of the evidence (\emph{response bias}). Two patterns stand out. First, richer channels help both models and humans, but not equally: audio adds reliable signal over text for each, yet only humans gain a further reliable increment from the visual channel, while the strongest models are flat from audio to audiovisual. They under-exploit the visible behavior humans read. Second, humans stay close to balanced across both classes in every modality, whereas models show a larger and more idiosyncratic class bias, most clearly in the text-only condition. Both patterns concern \emph{how} humans and models solve the task, rather than whether they succeed (Section~\ref{sec:analysis}).

\paragraph{Contributions.}
\begin{itemize}
\item We introduce \textsc{FriendBench}, a multimodal benchmark for inferring dyads' prior familiarity (familiar vs.\ strangers) from ice-breaker clips in which every dyad answers the same type of prompt, built on the Seamless Interaction dataset \citep{agrawal2025}.\footnote{The benchmark stimuli, human ratings, and model predictions are openly available at \url{https://huggingface.co/datasets/fluid-concepts/friend-bench}.}
\item We release a matched human-rater dataset for the task, covering 96 dyads in text, audio, and video, with roughly 90 raters per modality.
\item We evaluate 26 models from seven companies (OpenAI, Google, Anthropic, Alibaba, Mistral, Thinking Machines, and Meta), spanning proprietary and open-weight systems, across all three modalities.
\item We evaluate humans and models on the same stimuli under matched conditions. On accuracy, the best model and the human crowd are statistically indistinguishable in every modality. But equal accuracy is not human-like perception. The strongest models reach it by favoring ``stranger,'' while human raters stay balanced. Using signal detection theory, we characterize this as a difference in effective prior, not discrimination (Section~\ref{sec:analysis}).
\item We find a modality ordering both rater types obey only in part: text carries little signal for either and audio adds reliable discrimination, but the audiovisual channel gives \emph{humans} a further reliable gain while leaving the strongest models flat. Current models capture the vocal signal but under-exploit the visible behavior human observers read (Section~\ref{sec:modality}).
\end{itemize}

\section{Related Work}
\label{sec:related}

\paragraph{Thin-slice perception of familiarity.} Human observers form accurate judgments about people and relationships from very brief behavioral exposure. Ambady and Rosenthal's \citeyearpar{ambady1992} meta-analysis found that judgments from observations under five minutes (and often under 30 seconds) predicted objective outcomes at $r \approx .39$, with longer exposure adding little across studies.

Familiarity in particular is legible in thin slices, with friendship the best-studied case: observers tell friends from strangers from silent video \citep{latif2014} or brief audio \citep{bryant2020}, and cues such as inter-turn timing distinguish them \citep{templeton2023}. The closest precedent for our own task is the Interpersonal Perception Task \citep{costanzo1989}, which asks viewers to judge from a brief unscripted clip whether two people are strangers, friends, or relatives, and scores them against an objective key. Listeners manage this from the voice alone: \citet{montepare1988} found that people hearing one side of a phone call told an intimate male friend from a casual one at above-chance rates, and that the cue was paralinguistic rather than lexical. \citet{dunbar2022} show that relationship quality remains inferable from speech even after its lexical content is digitally removed. The relational signal need not come from what is said.

This motivates both our task and our 20-second window. Work that instead varies exposure within a single design does find modest gains from longer clips \citep{carney2007}. The two results fit together if a short slice already carries most of the available signal. Our 20 seconds is therefore a conservative choice rather than an accuracy-maximizing one.

\paragraph{How familiarity shapes interaction.} A separate literature shows where this signal comes from. Speakers tune greeting prosody to the person they address, and the form displays their stance toward the relationship \citep{pillet-shore2012}. Friends and strangers also coordinate differently. Friends produce more antiphonal laughter, meaning laughter that overlaps or immediately follows a partner's \citep{smoski2003}, and faster between-turn responses signal connection clearly enough for third parties to hear \citep{templeton2022}. Even function-word matching tracks relational outcomes \citep{ireland2011}. Familiarity also marks the transcript, since speakers use richer vocabulary and more complex grammar with familiar partners than with strangers \citep{luo2021}. These cues ride on timing, voice, and wording rather than topic, which our design isolates.

\paragraph{Recognizing relationships from behavior.} Machine analysis of nonverbal behavior for social inference has a long history \citep{vinciarelli2009}. One line of work predicts relationship \emph{type} from images or video via supervised classification: PISC \citep{li2017} labels images as intimate, non-intimate, or no-relation, PIPA \citep{sun2017} annotates sixteen fine-grained relations in photo albums, and more recent work classifies asymmetric relations from the temporal dynamics of a live interaction \citep{tang2026}.

A parallel line infers relationships from the \emph{semantic content} of dialogue: acoustic-lexical classifiers over phone calls \citep{katerenchuk2014} and language models over movie-script dialogue, from relation-classification datasets such as DDRel \citep{jia2021} to LLM evaluations where GPT-4o infers speaker relationships well above chance \citep{kim2026}. Both lines differ from our task in two ways: they infer relationship type rather than the presence or absence of prior familiarity, and they lean on appearance, scene, or semantic content (especially for scripted dialogue). We instead fix the conversational prompt, controlling the semantic content these methods lean on.

Two studies come closer to our question, addressing familiarity itself rather than relationship type. \citet{maynard1984} compared acquainted and unacquainted student dyads and found that unacquainted pairs build topics through pre-topical sequences that invite category self-identification, such as year in school or major, where acquainted pairs draw on shared knowledge instead. \citet{zhao2016} found that the temporal patterns predicting rapport in 30-second slices differ between friends and strangers.

\paragraph{Social reasoning in multimodal models.} Multimodal models are increasingly tested for social understanding: Social-IQ \citep{zadeh2019} poses questions about social videos, while SIV-Bench \citep{kong2026} and PIVOTSBench \citep{zhang2026a} probe reasoning about social scenes and fine-grained relations. HumanSense \citep{qin2026} is the closest to our setting. One of its subtasks asks a model to judge how well two people in a video know each other. Three things set our benchmark apart from these: (1) every dyad in our set answers the same type of prompt, so the topic itself cannot give the answer away; (2) our label is objective, recording whether a pair had actually met before rather than how close a viewer judges them; and (3) we gather matched human ratings in text, audio, and video for direct comparison.

Our stimuli come from the Seamless Interaction corpus \citep{agrawal2025}. Other dyadic corpora record acquaintance too, but treat it as metadata rather than as a label for prediction. UDIVA \citep{palmero2021} labels each pair known or unknown, and NoXi \citep{cafaro2017} rates how well partners know each other.

Comparison against a human panel is becoming standard practice. \citet{strachan2024} tested models and 1{,}907 people on classic theory-of-mind items, such as false belief and faux pas recognition, and traced an apparent model advantage on one of them to a response bias rather than to inference. Those items are written vignettes about mental states rather than perceptual judgments about a recorded interaction, so we take the design from that work rather than the task. Reported theory-of-mind success can also be brittle to small adversarial changes \citep{shapira2024}, which is one reason to test naturalistic stimuli. Such biases are well documented. A model's forced choice partly reflects a prior over answer labels separable from its ability to discriminate \citep{zhao2021}, and models can carry a systematic default answer that runs opposite to the human one \citep{braun2025}. We therefore report per-class recall and signal-detection statistics rather than accuracy alone.

\section{Methods}
\label{sec:methods}

\subsection{Benchmark Construction}
\label{sec:benchmark-construction}

Samples are drawn from the Seamless Interaction dataset \citep{agrawal2025}, restricted to naturalistic interactions from three recording sites (the dataset's \emph{vendor} field), each contributing a comparable share of dyads. We exclude the dataset's improvised interactions, in which participants are assigned a relationship and asked to act it out, since our aim is to measure this task on real, unscripted behavior rather than acted behavior. Because the benchmark is used only for zero-shot evaluation and never for training, we pool dyads across all of the dataset's predefined splits rather than restricting to one, maximizing the pool from which our quality filters and stratified design (below) can draw.

Seamless Interaction sessions include many different interaction types. We use only the Either-Or (EO) ice-breaker, a short task in which one participant poses a forced-choice hypothetical question (e.g., ``would you rather have the ability to fly or be invisible?'') and the dyad discusses their answer. 

We chose EO for three reasons: it standardizes the conversational context, since familiar and stranger pairs do the same thing and any signal must come from how a dyad responds rather than what it discusses; it is always the first interaction in a session, so every dyad is sampled from the same point in their interaction history; and its content carries little direct information about relationship status, so the task cannot be solved from semantic content alone. In free conversation, by contrast, status can leak through content: shared history and inside knowledge for familiars, getting-acquainted basics for strangers. A shared hypothetical question largely suppresses these cues.

An EO interaction, though centered on one either-or question, typically continues for several minutes. We sample one 20-second clip per dyad from this interaction, drawn from either its early or late portion (counterbalanced across relationship and recording site), with boundaries snapped ($\pm$5s) to the nearest turn start so clips do not begin or end mid-utterance. The rendered video stimulus retains the clip's audio track, so the video condition is audiovisual, with visible behavior together with speech, whereas the audio and text conditions each isolate a single channel. Both human raters and video models therefore receive sound as well as picture in the video condition.

Each dyad's relationship is a ground-truth label carried by the source dataset, recording whether the two participants were already acquainted (\emph{familiar}) or previously unacquainted (\emph{strangers}). The final set fully crosses recording site, relationship, and gender composition (same-gender/mixed-gender pair), with 8 dyads per cell, for 96 dyads that are balanced by construction (48 familiar, 48 strangers), and is participant-disjoint: no individual appears in more than one dyad. Crossing rules out recording site and gender composition as confounds (Section~\ref{sec:results}). Disjointness keeps each dyad an independent observation and guards against recognition leakage: because human raters see multiple stimuli in one session, a shared individual could be recognized from an earlier item, letting that recognition serve as the cue rather than the relationship itself.\footnote{This is not a risk for our models, which are evaluated zero-shot with no training process in which to learn identity.}

Interaction- and clip-level filters (minimum turns, in-window speaker balance, audio-track synchrony, interaction length, prompt completeness; see Table~\ref{tab:filters} in Appendix~\ref{app:quality-filters}) exclude clips that cannot support the task regardless of relationship, e.g., one participant speaking for only a few seconds or desynchronized audio tracks. An additional LLM-based audit checks that each interaction's recorded prompt label matches its content.

\subsection{Human Ratings}
\label{sec:human-ratings}

To establish baseline human performance, we collected human ratings of relationship type in a crowdsourced study. For each stimulus, raters made a forced choice: familiar or strangers. Before rating, each rater read a short instruction screen. It explained that they would read, listen to, or watch (depending on modality) a clip from a real conversation in which the two people were playing the ``Either Or'' ice-breaker game, then judge whether the pair had met before (\emph{familiar}) or were meeting for the first time (\emph{strangers}). The full instruction text is reproduced in Appendix~\ref{app:prompts}.

Raters were recruited via Prolific and completed the task on GORILLA. We ran three experiments, one per modality (text, audio, video), each with a panel of roughly 90 raters (94 text, 94 audio, 92 video). Prolific prescreening and quota matching ensured a gender-balanced pool of US nationals residing in the US, with no reported hearing difficulties or autism-spectrum diagnosis, and excluded anyone who had already participated in one of our other modality panels (see Appendix~\ref{app:study-details}).

Raters followed a planned-missing, 6-block incomplete design \citep{graham2006}, each rating one of six overlapping blocks and seeing 32 of the 96 dyads. Individual stimuli were in turn rated by 25--35 raters.
This design lets us model raters themselves, rather than averaging one rating per stimulus. We estimate each rater's own accuracy and response bias with crossed random effects. This estimate requires each rater to contribute enough trials to be more than a one-shot data point.

Each rater additionally completed the Social Information Processing (SP) subscale of the English-language Troms\o\ Social Intelligence Scale \citep[TSIS-E;][]{silvera2001,grieve2013} and a post-task self-report of which cues (visual, vocal, verbal, interactional) they attended to, for individual-differences analysis (Appendix~\ref{app:individual-differences}).

\subsection{Model Predictions}
\label{sec:model-predictions}

We evaluate 26 models from seven companies across text, audio, and video (Table~\ref{tab:full-models}), a fully-crossed design in which every model rates every stimulus. Unlike the human panels, no incomplete-block correction is needed for the model results. Each model receives the same prompt and response parsing. The model prompt is adapted directly from the human instruction text, so that both rater types are given the same task framing, the same ``Either Or'' description, and the same familiar/strangers definition. The only difference is the response mechanism: a forced-choice button for humans, a JSON label with a confidence rating for models. Models and prompts are further described in Appendix~\ref{app:model-config} and Appendix~\ref{app:prompts}, respectively.

A few audio models decline to answer on a share of trials (returning no usable familiar/strangers label) rather than committing to a forced choice. Because a refusal is not a wrong answer, we report each model's accuracy over its \emph{answered} (covered) trials and give per-model coverage alongside it (Table~\ref{tab:full-models}). Scoring refusals as errors would penalize a model for declining rather than for misjudging. Coverage is near-total except for three audio models (Section~\ref{sec:results}), and the top models in each modality answer every trial, so this choice does not affect the headline comparison.

\section{Benchmark Results}
\label{sec:results}

\begin{figure*}[t]
\centering
\includegraphics[width=\textwidth]{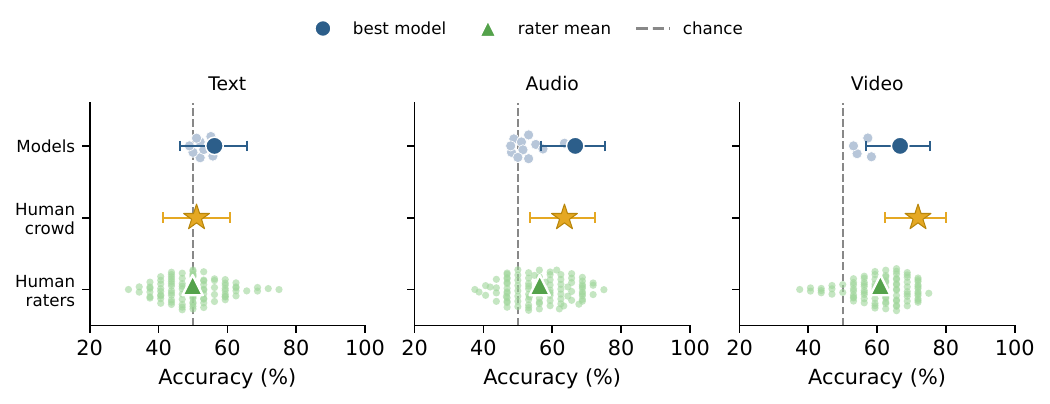}
\caption{Human vs.\ model accuracy by modality. Each rater point covers only the $\sim$32 trials seen by that rater, so the rater cloud is widened by sampling noise and its upper tail is not a band of expert raters (Section~\ref{sec:individual-differences}). Horizontal bars are Wilson 95\% CIs for the best model and crowd, not tests of human--model or cross-modality differences (Section~\ref*{sec:results}, Section~\ref*{sec:modality}).}

\label{fig:field}
\end{figure*}

Table~\ref{tab:comparison} summarizes, per modality, the average per-rater human, the human crowd (the per-stimulus majority vote of all raters who saw that stimulus), and the single best-performing model. For each it gives accuracy and, for the class-bias analysis in Section~\ref{sec:class-bias}, per-class recall and signal-detection statistics \citep{stanislaw1999}. Table~\ref{tab:full-models} gives the complete per-model breakdown with Wilson 95\% confidence intervals. Figure~\ref{fig:field} depicts the same comparison, plotting every model against the average individual human rater and the majority-vote crowd in each modality.

\begin{table}[t]
\centering
\small
\setlength{\tabcolsep}{4pt}
\begin{tabular}{l c @{\hskip 1.6em} c c r r}
\toprule
Rater & Acc. & Fam. & Str. & \multicolumn{1}{c}{$d'$} & \multicolumn{1}{c}{$c$} \\
\midrule
\multicolumn{6}{l}{\emph{Text}} \\
Human (indiv.) & 49.9 & 54.6 & 45.3 & 0.00 & $-0.12$ \\
Human (crowd)  & 51.0 & 59.6 & 42.6 & 0.05 & $-0.21$ \\
\texttt{gpt\_text\_mini} & 56.2 & 20.8 & 91.7 & 0.54 & $1.06$ \\
\midrule
\multicolumn{6}{l}{\emph{Audio}} \\
Human (indiv.) & 56.3\rlap{$^{**}$} & 57.2 & 55.5 & 0.32 & $-0.02$ \\
Human (crowd)  & 63.5\rlap{$^{*}$} & 64.6 & 62.5 & 0.68 & $-0.03$ \\
\texttt{gemini\_audio\_pro} & 66.7\rlap{$^{**}$} & 39.6 & 93.8 & 1.21 & $0.86$ \\
\midrule
\multicolumn{6}{l}{\emph{Video}} \\
Human (indiv.) & 60.9\rlap{$^{***}$} & 58.7 & 63.2 & 0.56 & $0.06$ \\
Human (crowd)  & 71.9\rlap{$^{***}$} & 70.2 & 74.5 & 1.16 & $0.06$ \\
\texttt{gemini\_video} & 66.7\rlap{$^{**}$} & 41.7 & 91.7 & 1.12 & $0.77$ \\
\bottomrule
\end{tabular}
\caption{Human raters (average individual and majority-vote crowd) vs.\ the best model per modality. Acc.\ is accuracy; Fam.\ and Str.\ per-class recall (all \%); $d'$ and $c$ as in Table~\ref{tab:full-models}. The named model in each block is that modality's best-performing (top-accuracy) system, scored over answered trials; human-indiv.\ accuracy is the per-rater mean (per-model Wilson CIs in Table~\ref{tab:full-models}). $^{*}/^{**}/^{***}$: above chance at $.05/.01/.001$, by exact binomial vs.\ 50\% (crowd, best model) or $95\%/99\%/99.9\%$ GLMM posterior HDI excluding 50\% (individual rows).}
\label{tab:comparison}
\end{table}

The best model and the human crowd trade the point-estimate lead across modalities. The model edges the crowd in text (by 5.2 points) and audio (by 3.2), while the crowd leads in video (by 5.2). Video is humans' strongest modality, and the one place the crowd, not a model, comes out ahead. None of these three gaps is statistically reliable, though. On the 96 paired dyads, an exact McNemar test \citep{dietterich1998,mcnemar1947} never approaches significance ($p>.4$ in every modality). Text is the weakest modality for humans: at chance individually and barely above it as a crowd. But the best text model is also not above chance (56.2\%, $p=.26$). Text is therefore less a model win than a modality where neither rater type finds a signal.

\begin{table*}[t]
\centering
\small
\begin{tabular}{lccrrrrr}
\toprule
Model & Acc. & 95\% CI & Cov. & Fam. & Str. & \multicolumn{1}{c}{$d'$} & \multicolumn{1}{c}{$c$} \\
\midrule
\multicolumn{8}{l}{\emph{Text}} \\
\mlogo{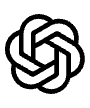}\texttt{gpt\_text\_mini}        & 56.2 & [46.3, 65.7] & 100 & 20.8 & 91.7 & 0.54 & $1.06$ \\
\mlogo{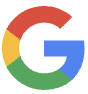}\texttt{gemini\_text\_thinking} & 55.8 & [45.8, 65.4] & 99  & 27.7 & 83.3 & 0.36 & $0.76$ \\
\mlogo{google.png}\texttt{gemini\_text}           & 55.2 & [45.3, 64.8] & 100 & 25.0 & 85.4 & 0.36 & $0.84$ \\
\mlogo{openai.png}\texttt{gpt\_text}              & 53.1 & [43.2, 62.8] & 100 & 29.2 & 77.1 & 0.19 & $0.63$ \\
\mlogo{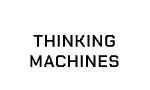}\texttt{inkling\_text}          & 52.1 & [42.2, 61.8] & 100 & 16.7 & 87.5 & 0.17 & $1.03$ \\
\mlogo{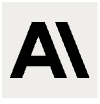}\texttt{claude\_text}           & 52.1 & [42.2, 61.8] & 100 & 14.6 & 89.6 & 0.19 & $1.12$ \\
\mlogo{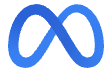}\texttt{muse\_spark\_text\_high}& 51.0 & [41.2, 60.8] & 100 & 8.3  & 93.8 & 0.14 & $1.40$ \\
\mlogo{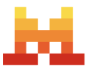}\texttt{mistral\_text}          & 50.0 & [40.2, 59.8] & 100 & 87.5 & 12.5 & 0.00 & $-1.11$ \\
\mlogo{meta.png}\texttt{muse\_spark\_text}      & 49.0 & [39.2, 58.8] & 100 & 10.4 & 87.5 & $-0.10$ & $1.16$ \\
\midrule
\multicolumn{8}{l}{\emph{Audio}} \\
\mlogo{google.png}\texttt{gemini\_audio\_pro}      & 66.7\rlap{$^{**}$} & [56.8, 75.3] & 100 & 39.6 & 93.8 & 1.21 & $0.86$ \\
\mlogo{google.png}\texttt{gemini\_audio}           & 63.5\rlap{$^{*}$}  & [53.6, 72.5] & 100 & 81.2 & 45.8 & 0.76 & $-0.48$ \\
\mlogo{meta.png}\texttt{muse\_spark\_audio}      & 57.3 & [47.3, 66.7] & 100 & 20.8 & 93.8 & 0.67 & $1.13$ \\
\mlogo{google.png}\texttt{gemini\_audio\_thinking} & 55.2 & [45.3, 64.8] & 100 & 64.6 & 45.8 & 0.26 & $-0.23$ \\
\mlogo{meta.png}\texttt{muse\_spark\_audio\_high}& 53.1 & [43.2, 62.8] & 100 & 18.8 & 87.5 & 0.25 & $0.99$ \\
\mlogo{openai.png}\texttt{gpt\_audio\_1\_5}        & 53.1 & [43.2, 62.8] & 100 & 87.5 & 18.8 & 0.25 & $-0.99$ \\
\mlogo{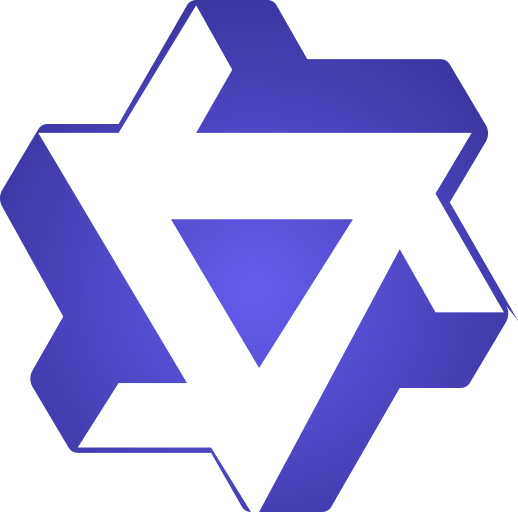}\texttt{qwen\_audio}             & 51.5 & [39.8, 62.9] & 71  & 0.0 & 100.0 & 0.02 & $2.19$ \\
\mlogo{thinking-machines-small.png}\texttt{inkling\_audio}          & 51.0 & [41.2, 60.8] & 100 & 41.7 & 60.4 & 0.05 & $0.23$ \\
\mlogo{mistral.png}\texttt{voxtral\_audio}          & 50.0 & [40.2, 59.8] & 100 & 100.0 & 0.0 & 0.00 & $-2.32$ \\
\mlogo{openai.png}\texttt{gpt\_audio}              & 48.8 & [34.6, 63.2] & 45  & 100.0 & 4.3 & 0.45 & $-1.76$ \\
\mlogo{openai.png}\texttt{gpt\_audio\_mini}        & 48.1 & [30.7, 66.0] & 28  & 25.0 & 81.8 & 0.18 & $0.72$ \\
\mlogo{qwen.png}\texttt{qwen\_omni}              & 47.9 & [38.2, 57.8] & 100 & 16.7 & 79.2 & $-0.15$ & $0.87$ \\
\midrule
\multicolumn{8}{l}{\emph{Video}} \\
\mlogo{google.png}\texttt{gemini\_video}          & 66.7\rlap{$^{**}$} & [56.8, 75.3] & 100 & 41.7 & 91.7 & 1.12 & $0.77$ \\
\mlogo{google.png}\texttt{gemini\_video\_thinking}& 58.3 & [48.3, 67.7] & 100 & 27.1 & 89.6 & 0.62 & $0.91$ \\
\mlogo{google.png}\texttt{gemini\_video\_pro}     & 57.3 & [47.3, 66.7] & 100 & 18.8 & 95.8 & 0.77 & $1.25$ \\
\mlogo{meta.png}\texttt{muse\_spark\_video}     & 54.2 & [44.2, 63.8] & 100 & 14.6 & 93.8 & 0.44 & $1.24$ \\
\mlogo{meta.png}\texttt{muse\_spark\_video\_high}& 53.1 & [43.2, 62.8] & 100 & 10.4 & 95.8 & 0.42 & $1.42$ \\
\bottomrule
\end{tabular}
\caption{Per-model results, sorted within modality. Acc.\ is accuracy (\%) over answered trials and CI is its Wilson 95\% confidence interval; Cov.\ is coverage (\% answered rather than declined). Fam.\ and Str.\ are per-class recall (\%): share of familiar and of stranger dyads correctly labeled. $d'$ and criterion $c$ treat ``familiar'' as the signal class (log-linear corrected; \citealp{hautus1995}), so $c>0$ indicates a bias toward ``stranger'' and $c<0$ toward ``familiar.'' $^{**}$/$^{*}$: accuracy above chance at $p<.01$/$p<.05$ (exact binomial vs.\ 50\%).}
\label{tab:full-models}
\end{table*}

Only three models are individually above chance at this sample size, all from Google: \texttt{gemini\_audio\_pro} and \texttt{gemini\_audio} in audio, and \texttt{gemini\_video} in video. No text model clears chance. Apart from the two top-accuracy systems, the best remaining model in each modality reaches only 56.2\% (text, \texttt{gpt\_text\_mini}), 63.5\% (audio, \texttt{gemini\_audio}), and 58.3\% (video, \texttt{gemini\_video\_thinking}). Most models cluster near chance on raw accuracy. The per-class recall and criterion columns already hint at the analysis in Section~\ref{sec:class-bias}: several models carry very large response biases regardless of accuracy. In audio the bias is large but inconsistent in direction. \texttt{voxtral\_audio} labels every dyad ``familiar'' (100\%/0\%, $c=-2.32$), while \texttt{qwen\_audio} does the opposite (0\%/100\%, $c=2.19$). In text most models strongly favor ``stranger,'' with \texttt{muse\_spark\_text\_high} recovering only 8.3\% of familiar pairs ($c=1.40$). A near-chance accuracy can thus conceal a strongly one-sided response pattern rather than balanced guessing.

Three audio models also decline a substantial share of trials: \texttt{gpt\_audio} answers only 45\%, \texttt{gpt\_audio\_mini} 28\%, and \texttt{qwen\_audio} 71\%. They sit at chance on the trials they do answer, so their behavior is better summarized as declining-plus-guessing.

\section{Performance Analysis}
\label{sec:analysis}

\subsection{Class Bias: Humans vs.\ Models}
\label{sec:class-bias}

Humans and the best-performing models are close to parity on accuracy (Section~\ref{sec:results}), but do they reach it the same way? To separate \emph{discrimination} from \emph{response bias}, we read the per-class recall and signal-detection columns of Table~\ref{tab:comparison}, treating ``familiar'' as the signal class: $d'$ measures how well a rater tells the classes apart, and the criterion $c$ how far its decision rule favors one answer ($c>0$ toward ``stranger,'' $c<0$ toward ``familiar''). Two patterns stand out. Humans stay near-balanced: the criterion is within $0.23$ of zero in every modality, and per-class recall gaps never exceed 10 points for pooled raters. The best text and audio models instead reach their accuracy largely through response bias: \texttt{gemini\_audio\_pro} labels 93.8\% of strangers correctly but recovers only 39.6\% of familiar pairs ($c=0.86$), and the best text model, \texttt{gpt\_text\_mini}, is more extreme (91.7\% vs.\ 20.8\%, $c=1.06$). Across the roster this is the rule: in text almost every model favors ``stranger'' by 50--85 recall points, and in audio the bias is large but inconsistent in direction (Table~\ref{tab:full-models}).

The gap holds even in video, where models come closest to humans. There the best model (\texttt{gemini\_video}) still favors ``stranger'' ($c=0.77$; 91.7\% of strangers but only 41.7\% of familiar pairs), while the human crowd stays balanced ($c=0.06$) at slightly higher discrimination ($d'=1.16$ vs.\ $1.12$). The two reach comparable $d'$, but only the crowd does so without skewing its decision rule. Models thus fail to recreate the human balanced-recall pattern in \emph{any} modality, including their strongest: several of the highest-accuracy systems (Table~\ref{tab:full-models}) buy that accuracy with a skewed criterion rather than sharper discrimination.

\subsection{Wisdom of the Crowd}
\label{sec:crowd}

\begin{figure}[t]
\centering
\includegraphics[width=\columnwidth]{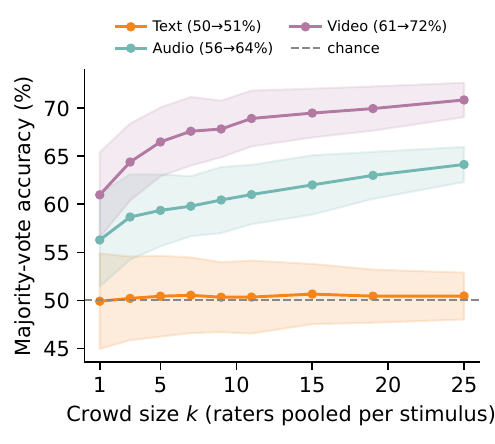}
\caption{Wisdom of the crowd. Majority-vote accuracy as a function of human crowd size $k$, with raters sampled per stimulus from that stimulus's own rater panel (mean $\pm$ SD over 300 bootstrap draws). Legend gives single-rater~$\to$~full-panel crowd accuracy per modality.}
\label{fig:crowd}
\end{figure}

Relative to the average single rater, pooling human raters via majority vote barely moves accuracy in text (49.9\%$\to$51.0\%, $+$1.1 points) but raises it substantially in audio (56.3\%$\to$63.5\%, $+$7.2 points) and video (60.9\%$\to$71.9\%, $+$11.0 points). In each pair the first number is the mean single-rater accuracy and the second the full-panel majority vote. Figure~\ref{fig:crowd} shows the full crowd-growth curves: majority-vote accuracy rises with crowd size $k$ in audio and video but stays flat near chance in text. The audio and video gains therefore reflect real signal, not an artifact of pooling.

\subsection{Modality Comparison}
\label{sec:modality}

Text is the weakest modality for both humans and models, and richer channels help both, though not to the same degree, as we show below. To test the modality ordering while accounting for the repeated-measures structure of the human data, we fit a Bayesian crossed random-effects logistic model \citep{gelman2014} to per-trial human correctness (Bernoulli likelihood, logit link), with a fixed effect of modality and crossed random intercepts for rater and for dyad (Section~\ref{sec:human-ratings}). This is the design the block structure was built to support. We estimate it with the \texttt{bambi} interface to PyMC \citep{capretto2022}, using default weakly-informative, data-scaled priors (wide normal priors on the fixed effects and half-normal hyperpriors on the random-effect standard deviations), and sample with NUTS (4 chains, 1{,}000 warmup and 1{,}000 post-warmup draws each; all $\hat{R}\approx1.00$). We report population-averaged (marginal) accuracies and contrasts as posterior means with 95\% highest-density intervals (HDIs, \citealp{makowski2019}).

The marginal accuracies confirm the raw pattern: text $50.0\%$ (95\% HDI $[46.7, 53.6]$), audio $56.6\%$ $[52.2, 60.6]$, video $60.9\%$ $[56.9, 64.9]$. The pairwise contrasts are decisive: audio exceeds text by $6.5$ points (HDI $[4.1, 8.9]$), video exceeds audio by $4.5$ points ($[1.9, 6.8]$), and video exceeds text by $10.9$ points ($[8.5, 13.3]$), each with posterior $P>0$ of at least $0.999$. Text alone is statistically indistinguishable from chance (posterior probability of exceeding 50\% only $0.50$). Because the video condition is audiovisual (Section~\ref{sec:benchmark-construction}), its edge over audio reflects the \emph{added} value of visible behavior on top of speech, not vision in isolation. Video is best read as an audiovisual upper bound rather than a vision-only channel.

The model side matches the human floor but not the human ceiling. The best model climbs from text ($56.2\%$) to audio ($66.7\%$) but gains nothing from the audiovisual channel ($66.7\%$), and among models evaluated in both audio and video only Gemini~Flash improves ($63.5\%\to66.7\%$) while Gemini~Pro ($66.7\%\to57.3\%$) and Muse~Spark ($57.3\%\to54.2\%$) decline. (Cross-modality model means rise monotonically: $52.7$, $53.9$, $57.9\%$ for text, audio, video. But the video roster is small and skewed toward stronger companies, so we read the best-model and within-model trajectories rather than the pooled mean.) Humans reliably convert visible behavior into accuracy on top of speech, but the strongest models do not. They capture the vocal signal but under-exploit the visual channel. That channel is where a benchmark of \emph{multimodal} social perception most expects a model to gain.

That text is the hardest modality is consistent with the benchmark's design rather than a defect of it. The EO ice-breaker was chosen precisely so that transcript content carries little direct information about relationship status (Section~\ref{sec:benchmark-construction}). The weak transcript performance of both humans and models is the expected consequence of that choice. It also helps explain why response bias is most visible in text: with little signal available, a rater's responses reflect its bias more than the stimulus.

\subsection{Rater Individual Differences}
\label{sec:individual-differences}

Because each rater contributes many trials by design (Section~\ref{sec:human-ratings}), we can estimate individual accuracy and relate it to rater traits. Trait social intelligence (SP subscale; reliable in every panel, $\alpha=0.88$--$0.92$) is essentially uncorrelated with accuracy in the two modalities where the task is doable: audio $r=0.03$ ($p=.78$) and video $r=-0.12$ ($p=.24$). The only significant association is in text ($r=0.29$, $p=.004$), the modality where average accuracy is at chance, so we read it with caution. Self-reported cue use is similarly flat: raters most often report attending to interactional cues (rapport, responsiveness), but cue-use scores rarely predict accuracy (Appendix~\ref{app:individual-differences}).

The individual-rater cloud in Figure~\ref{fig:field} is correspondingly wide, with the best raters near 75\% in every modality. Its spread should not be read as a stable band of expert observers, though. Each rater's accuracy comes from only the $\sim$32 trials they saw, so the upper tail is close to what sampling noise alone would produce. The best model therefore sits within the human distribution, not above it, even as almost no individual beats the aggregated crowd (1 of 92 in video). The crossed random-effects model makes the point directly: the dyad random-intercept SD ($0.67$--$0.95$ across modalities, logit scale) is six to seven times the rater SD ($0.10$--$0.14$). Who the rater is thus matters far less than which dyad they were rating.

\subsection{Dyad Difficulty}
\label{sec:dyad-difficulty}

Treating the per-dyad random intercepts from the crossed random-effects model as Rasch-style item-easiness parameters \citep{rasch1960,deboeck2004}, we find that difficulty is largely a property of the conversation, not of the modality through which it is observed. Estimated dyad easiness correlates positively across all modality pairs (text--audio $r=0.52$, text--video $r=0.33$, audio--video $r=0.61$; all $p<.01$): a dyad that is hard to read from one modality tends to be hard from the others as well. A handful of dyads sit below chance in every modality, acting as systematic ``lures'' that most raters misread alike (Appendix~\ref{app:dyad-difficulty}).

Humans and models tend to find the same dyads hard, though how strongly depends on how it is measured. Correlating per-dyad human-crowd accuracy with per-dyad accuracy pooled over all models yields $r=0.56$ (audio) and $r=0.34$ (video), both significant ($p<.01$; Appendix Figure~\ref{fig:difficulty}), but only $r=0.20$ in text ($p=.046$). This raw correlation understates the shared difficulty. It blends two things: how hard a dyad is to read, and which class it belongs to. Humans stay balanced while the models favor ``stranger'' (Section~\ref{sec:class-bias}), so the two rater types tend to miss different classes. Isolating difficulty from this class split, by correlating within each true class, brings the agreement out clearly: $r=0.61$ (text), $0.58$ (audio), and $0.44$ (video), all $p<.001$. So a substantial part of what makes a pair legible or illegible is a property of the interaction shared across observers, even though the two reach their answers by different rules.

\section{Discussion}
\label{sec:discussion}

For a benchmark of multimodal social perception, the clearest result is about the modalities themselves. Text carries little relational signal for anyone, by design: the shared ice-breaker prompt suppresses lexical content (Section~\ref{sec:benchmark-construction}). Richer channels help both humans and models, but asymmetrically. Humans improve reliably at every step, from text to audio to audiovisual (Section~\ref{sec:modality}), gaining a credible increment from visible behavior on top of speech. The strongest models capture the audio gain but not the visual one: their top accuracy is identical in audio and audiovisual ($66.7\%$). The sharpest human--model difference is thus not whether models can read relationships but whether they exploit the visual channel as humans do. This gap has applied stakes. The companion agents, meeting assistants, and care monitors that motivate the task must read social situations from behavior, and the visible cues humans exploit on top of speech are exactly what current models miss.

On overall accuracy, the best models have drawn level with the human crowd. In every modality the two are statistically indistinguishable. Point estimates put the model slightly ahead in text and audio and the crowd slightly ahead in video (Table~\ref{tab:comparison}), but no gap survives a paired McNemar test ($p>.4$ throughout; Section~\ref{sec:results}).

It may seem surprising that human accuracy is not higher, since people are known to read relationships well from brief exposure (Section~\ref{sec:related}). That prior work sets a realistic bar rather than a high one. Its accuracy effects are moderate in size, near $r=.39$, and not close to ceiling. Our task is also hard by design. Each judgment rests on a single 20-second clip, the response is a forced choice, and the balanced set removes any base-rate cue that a rater could use as a shortcut. The human signal is nonetheless real. The crowd reaches about 72\% in video, and the best raters approach 75\% in every modality (Section~\ref{sec:individual-differences}).

The difference is in how humans and models use their two answers. Discrimination ($d'$) measures how well a rater tells the classes apart. Response bias (criterion $c$; Table~\ref{tab:comparison}) measures how far it favors one answer. Human raters stay balanced across the two answers in every modality. The strongest models instead favor ``stranger,'' giving that answer more often regardless of the pair. Video makes this clearest. There the best model tells the classes apart as well as the crowd ($d'\approx1.1$), but it reaches that accuracy by favoring ``stranger'' where the crowd stays balanced.

In signal-detection terms, this is a difference in \emph{effective prior}: the models behave as though strangers were the more common answer and demand more evidence before saying ``familiar.'' Humans behave as though the classes were equally likely, which is true in our balanced set. This describes the response distribution, not a mechanism, and the prior is inferred from behavior rather than verified. A pure criterion shift is correctable: recentering to the known base rate would raise accuracy with no gain in discrimination. So a biased model's raw accuracy can understate its discrimination.

We cannot say from these data why models adopt this prior. One simple explanation is that strangers are more often the correct answer in the world, so favoring that answer would be well calibrated rather than mistaken. This does not fit our results well. The size of the bias, and even its direction, varies widely from model to model (Table~\ref{tab:full-models}), which is not what a single shared world prior would produce. We therefore treat this skew as a property of each model's decision rule, not as evidence that models have learned the true base rate.

At the same time, humans and models are not perceiving unrelated things. Dyad difficulty is partially shared and partially transfers across modalities (Section~\ref{sec:dyad-difficulty}), so part of what makes a pair legible or illegible is a property of the interaction itself, not of the observer or channel. The dissociation is therefore specific: the two agree on \emph{which} dyads are hard but diverge on the \emph{decision rule} they apply. Methodologically, these results argue for evaluating social-perception models with more than a single accuracy number: per-class recall, signal-detection statistics, and item-level difficulty each revealed structure that accuracy alone obscured, and are cheap to report once the underlying predictions are available.

Overall, the strongest models now match the human crowd on accuracy, but not on how they reach it. They find the same conversations hard. Yet they systematically favor ``stranger,'' where human raters stay balanced, and they do not convert the visual channel into accuracy. This benchmark is built to track whether models can be brought to read relationships as humans do, balanced across classes and drawing on visible behavior rather than speech alone.

\section*{Limitations}
\label{sec:limitations}

This paper reports only the ice-breaker (first-interaction) task from the Seamless Interaction dataset. Findings about modality strength and class bias may not generalize to conversations later in a relationship or session, or to different prompt structures. All familiar subtypes (friends, family, romantic partner, coworkers, familiar\_other) are collapsed into a single ``familiar'' class, because the balanced design does not have enough dyads per subtype to power a multi-class analysis. A six-class relationship-type task is defined in the broader project but is out of scope here, so any within-familiar heterogeneity is invisible to this binary framing.

Only two-person interactions are studied, and relationship inference in larger groups may draw on cues not captured here. The balanced evaluation set is also modest in size (96 dyads, one clip each), so per-model accuracy intervals are wide and only the strongest models clear chance individually. The class-bias and difficulty patterns we emphasize are more robust than any single model's rank.

Our human baseline is drawn entirely from US-resident raters (Section~\ref{sec:human-ratings}). Relationship-perception cues can be culturally specific, so this baseline may not represent human performance in other populations. Finally, even the ``naturalistic'' subset was recorded in a fixed-camera motion-capture studio, with wired lapel microphones and a posed ice-breaker prompt. This is not truly in-the-wild interaction, so findings may not transfer to less controlled contexts such as phone video or casual settings.

We also do not identify which cues drive these judgments. Our measures show whether each rater type separates the two classes and how it biases its answers, but not what in the clip carries the signal, whether that is prosody, word choice, facial or bodily movement, or the timing of the exchange. We do collect rater self-reports of the cues people say they attended to (Section~\ref{sec:individual-differences}), but these rarely predict accuracy and do not tell us which cues actually matter, for humans or for models. Identifying them, for example through controlled removal of parts of the input or through probing of model representations, is useful future work.

Our evaluation set is also balanced 50/50 between familiar and stranger dyads by construction, which makes accuracy a clean measure of discrimination but base-rate-specific. A reader might ask whether the models' skew toward ``stranger'' is not miscalibration but a well-calibrated prior for a world in which two people recorded together are more often strangers, penalized only by our artificial balance. This does not threaten our central claim. That claim is the \emph{difference} in criterion between humans and models on matched stimuli. Neither rater type was told the base rate, so this difference does not depend on the true base rate, and neither does $d'$, which is base-rate-invariant. It does mean we measure discrimination, not deployment calibration. We do not read the balanced accuracy as a deployment estimate. As noted in Section~\ref{sec:discussion}, this bias is idiosyncratic across models, which argues against calibration to any single real-world base rate. Measuring calibration against realistic base rates would complement the discrimination-focused evaluation we report here.

Each model is also evaluated under a single prompt, adapted from the human instructions (Section~\ref{sec:model-predictions}). Because model responses can be sensitive to prompt wording, the exact biases in Table~\ref{tab:full-models} may shift under other phrasings. A prompt-robustness sweep is worthwhile future work. The human--model comparison holds the task framing fixed for both rater types, so it is less exposed to this concern than any single model's bias estimate read on its own.

\section*{Ethics Statement}
\label{sec:ethics}

\paragraph{Informed consent and compensation.} Rater recruitment and prescreening are described in Section~\ref{sec:human-ratings} and Appendix~\ref{app:study-details}. Before viewing any stimulus, each rater read a description of the study and affirmatively agreed to three statements: that they had read and understood the information, that they were free to withdraw at any time, by closing the browser tab, without giving a reason, and that they agreed to take part. Participation was voluntary and could be ended at any point without penalty. Sessions took roughly 15--30 minutes depending on modality, and raters were compensated at an effective rate of approximately \$10--12 per hour, above Prolific's fair-pay guidance. The task was minimal-risk: raters viewed short, benign clips of consenting conversation partners and answered non-sensitive perceptual and self-report questions.

\paragraph{Privacy and data minimization.} We collected no personally identifying information from raters. Beyond the task responses and self-report questionnaires analyzed here, we recorded only the coarse attributes used for prescreening and quota matching (Section~\ref{sec:human-ratings}), keyed to a pseudonymous Prolific identifier we do not link to any real-world identity. All released rater data are de-identified.

\paragraph{Institutional review.} This research was conducted outside a university setting and was not reviewed by an institutional review board. Because it collected no personally identifying information from raters and involved only minimal-risk procedures, it did not fall under IRB oversight. We nonetheless followed standard human-subjects safeguards: voluntary informed consent, the right to withdraw without penalty, fair compensation, minimal-risk stimuli, and data de-identification.

\paragraph{Stimulus source.} All interaction clips are drawn from the publicly released Seamless Interaction dataset \citep{agrawal2025}, whose participants consented to the recording and research use of their audio and video. We use these data in accordance with the dataset's CC-BY-NC 4.0 license (attribution, non-commercial use only): the rendered benchmark clips are redistributed under the same CC-BY-NC 4.0 terms, with attribution to the Seamless Interaction dataset. We do not attempt to re-identify or contact any recorded individual.

\paragraph{Intended use and risks.} The benchmark is intended for evaluating and auditing the social-perceptual behavior of models, including the response biases we document (Section~\ref{sec:class-bias}). Inferring familiarity from behavior could in principle support surveillance or profiling. We release the benchmark to enable research on and scrutiny of such capabilities, not their deployment. Given the modest accuracy and pronounced, idiosyncratic biases we observe (Section~\ref{sec:results}), we caution against using these models or this task to make consequential judgments about real individuals.

\bibliography{references}

\clearpage

\appendix

\renewcommand{\thetable}{A\arabic{table}}
\renewcommand{\thefigure}{A\arabic{figure}}
\setcounter{table}{0}
\setcounter{figure}{0}


\section{Benchmark Quality Filters}
\label{app:quality-filters}


Every candidate interaction and clip must pass the automatic quality filters in Table~\ref{tab:filters} before entering the stratified selection pool (Section~\ref{sec:benchmark-construction}). Interaction-level filters (F1, F4, F5) gate the whole source interaction; clip-level filters (F2, F3) gate the specific 20-second window sampled from it. Clips are 20\,s long with boundaries snapped within $\pm5$\,s to the nearest turn start. These filters remove clips that cannot support the task regardless of relationship; they are followed by the LLM-based prompt-adherence audit (Section~\ref{sec:benchmark-construction}), which verifies that each interaction's recorded prompt label matches its actual content.

\begin{table*}[t]
\centering
\small
\begin{tabular}{@{}llp{0.78\linewidth}@{}}
\toprule
ID & Level & Criterion \\
\midrule
F1 & Interaction & Both recorded EO prompts are free of placeholder/boilerplate text (e.g., ``you will each see the same prompt,'' ``given different lists of sentences'') that signals the true question was not logged. \\
F2 & Clip & $\geq 4$ merged conversational turns within the clip window. \\
F3 & Clip & Each speaker contributes $\geq 3$\,s of speech within the clip window. \\
F4 & Interaction & $< 35\%$ of cross-track turn-start pairs coincide within $0.2$\,s, rejecting duplicated or desynchronized audio tracks. \\
F5 & Interaction & $\geq 90$\,s of speech in the interaction. \\
\bottomrule
\end{tabular}
\caption{Automatic quality filters applied during clip generation and the EO interaction audit. Interaction-level filters gate the source interaction; clip-level filters gate the sampled 20\,s window. Thresholds are the generator/auditor constants (\texttt{scripts/generate\_samples\_rapport.py}, \texttt{scripts/audit\_eo\_interactions.py}).}
\label{tab:filters}
\end{table*}

\section{Model Inference Configuration}
\label{app:model-config}


Table~\ref{tab:model-config} lists the inference configuration for each of the 26 models. All models received the same prompt construction and response parsing (Section~\ref{sec:model-predictions}); the settings below cover per-model decoding and, where applicable, reasoning parameters. Cloud models were queried at temperature $0$ with a fixed seed ($42$) wherever the API exposed them; the three locally-run open-weight models (Qwen2-Audio, Qwen2.5-Omni, Voxtral) used greedy decoding (\texttt{do\_sample=False}), which is deterministic without a seed. Answer length was capped at 150--300 tokens for non-reasoning models, while reasoning models received a larger combined answer-plus-reasoning budget (8{,}000 tokens for the Gemini thinking and pro variants; 4{,}000 for Inkling and Muse~Spark) to avoid truncating the response.

\begin{table*}[t]
\centering
\small
\begin{tabular}{lllll}
\toprule
Model & API / model ID & Temp. & Seed & Reasoning \\
\midrule
\multicolumn{5}{l}{\emph{Text}} \\
\mlogo{openai.png}\texttt{gpt\_text}              & \texttt{gpt-4o}                         & $0$      & $42$      & --- \\
\mlogo{openai.png}\texttt{gpt\_text\_mini}        & \texttt{gpt-4o-mini}                    & $0$      & $42$      & --- \\
\mlogo{google.png}\texttt{gemini\_text}           & \texttt{gemini-3.5-flash}               & $0$      & $42$      & off \\
\mlogo{google.png}\texttt{gemini\_text\_thinking} & \texttt{gemini-3.5-flash}               & $0$      & $42$      & dynamic \\
\mlogo{anthropic.png}\texttt{claude\_text}           & \texttt{claude-opus-4-8}                & ---$^{a}$ & ---$^{a}$ & off$^{a}$ \\
\mlogo{mistral.png}\texttt{mistral\_text}          & \texttt{mistral-large-latest}           & $0$      & $42$      & --- \\
\mlogo{thinking-machines-small.png}\texttt{inkling\_text}          & \texttt{thinkingmachines/Inkling}       & $0$      & $42^{b}$  & on \\
\mlogo{meta.png}\texttt{muse\_spark\_text}      & \texttt{muse-spark-1.1}                 & ---$^{c}$ & ---$^{c}$ & minimal \\
\mlogo{meta.png}\texttt{muse\_spark\_text\_high}& \texttt{muse-spark-1.1}                 & ---$^{c}$ & ---$^{c}$ & high \\
\midrule
\multicolumn{5}{l}{\emph{Audio}} \\
\mlogo{openai.png}\texttt{gpt\_audio}              & \texttt{gpt-audio}                      & $0$      & $42$      & --- \\
\mlogo{openai.png}\texttt{gpt\_audio\_1\_5}        & \texttt{gpt-audio-1.5}                  & $0$      & $42$      & --- \\
\mlogo{openai.png}\texttt{gpt\_audio\_mini}        & \texttt{gpt-audio-mini}                 & $0$      & $42$      & --- \\
\mlogo{google.png}\texttt{gemini\_audio}           & \texttt{gemini-3.5-flash}               & $0$      & $42$      & off \\
\mlogo{google.png}\texttt{gemini\_audio\_pro}      & \texttt{gemini-pro-latest}              & $0$      & $42$      & dynamic$^{d}$ \\
\mlogo{google.png}\texttt{gemini\_audio\_thinking} & \texttt{gemini-3.5-flash}               & $0$      & $42$      & dynamic \\
\mlogo{qwen.png}\texttt{qwen\_audio}             & \texttt{Qwen/Qwen2-Audio-7B-Instruct}   & greedy   & ---       & --- \\
\mlogo{qwen.png}\texttt{qwen\_omni}              & \texttt{Qwen/Qwen2.5-Omni-7B}           & greedy   & ---       & --- \\
\mlogo{mistral.png}\texttt{voxtral\_audio}          & \texttt{mistralai/Voxtral-Mini-3B-2507} & greedy   & ---       & --- \\
\mlogo{thinking-machines-small.png}\texttt{inkling\_audio}          & \texttt{thinkingmachines/Inkling}       & $0$      & $42^{b}$  & on \\
\mlogo{meta.png}\texttt{muse\_spark\_audio}      & \texttt{muse-spark-1.1}                 & ---$^{c}$ & ---$^{c}$ & minimal \\
\mlogo{meta.png}\texttt{muse\_spark\_audio\_high}& \texttt{muse-spark-1.1}                 & ---$^{c}$ & ---$^{c}$ & high \\
\midrule
\multicolumn{5}{l}{\emph{Video}} \\
\mlogo{google.png}\texttt{gemini\_video}           & \texttt{gemini-3.5-flash}               & $0$      & $42$      & off \\
\mlogo{google.png}\texttt{gemini\_video\_pro}      & \texttt{gemini-pro-latest}              & $0$      & $42$      & dynamic$^{d}$ \\
\mlogo{google.png}\texttt{gemini\_video\_thinking} & \texttt{gemini-3.5-flash}               & $0$      & $42$      & dynamic \\
\mlogo{meta.png}\texttt{muse\_spark\_video}      & \texttt{muse-spark-1.1}                 & ---$^{c}$ & ---$^{c}$ & minimal \\
\mlogo{meta.png}\texttt{muse\_spark\_video\_high}& \texttt{muse-spark-1.1}                 & ---$^{c}$ & ---$^{c}$ & high \\
\bottomrule
\end{tabular}
\caption{Per-model inference configuration (26 models, seven companies; roster matches Table~\ref{tab:full-models}). \emph{Temp.}\ and \emph{Seed} are the decoding temperature and random seed passed to the API; \texttt{greedy} marks the locally-run open-weight models decoded with \texttt{do\_sample=False} (deterministic, no seed). \emph{Reasoning}: ``---'' = non-reasoning model; ``off'' = reasoning-capable but disabled (Gemini \texttt{thinking\_budget}${=}0$); ``dynamic'' = model sets its own reasoning depth (\texttt{thinking\_budget}${=}{-}1$); ``on'' = reasoning always on with no depth control; ``minimal''/``high'' = named reasoning-effort tier. Two Gemini IDs are aliases; at run time (July 2026) \texttt{gemini-pro-latest} resolved to \texttt{gemini-3.1-pro-preview}, and \texttt{gemini-3.5-flash} was pinned explicitly rather than \texttt{gemini-flash-latest} (which then resolved to \texttt{gemini-3.6-flash}). $^{a}$Claude rejects \texttt{temperature}/\texttt{top\_p} and exposes no seed; extended thinking was not enabled. $^{b}$Inkling accepts a seed but the API echoed \texttt{null}, so determinism is best-effort. $^{c}$Muse~Spark exposes no temperature or seed control, and its reasoning depth is non-deterministic run-to-run. $^{d}$\texttt{gemini-pro-latest} cannot disable thinking and always reasons dynamically.}
\label{tab:model-config}
\end{table*}

\section{Task Instructions and Prompts}
\label{app:prompts}


Both human raters and models were given the same task framing, ``Either Or'' game description, and familiar/strangers definition, differing only in how a response was collected. Human raters selected an answer with a button; models were additionally instructed to emit a JSON object with a label and confidence. Modality-specific slots (shown here in brackets, Text/Audio/Video order) were filled per panel or per clip.

\paragraph{Human rater instructions.}
\begin{small}
\begin{verbatim}
You will [read / listen to / watch] a series
of [transcripts / audio clips / video clips]
taken from real conversations between two
people.

These people are playing a game called
"Either Or," where they discuss whether they
would prefer one thing -- for example, the
ability to fly -- or an alternative, such as
the ability to breathe underwater.

Your job is to pay close attention to each
conversation and determine whether the two
people have met before -- meaning they are
familiar, such as friends, family members,
coworkers, or romantic partners -- or are
meeting for the first time in this
interaction -- meaning they are strangers.
\end{verbatim}
\end{small}
\noindent Raters then made a forced choice: \emph{familiar} or \emph{strangers}.

\paragraph{Model prompt.}
\begin{small}
\begin{verbatim}
[Read / Listen to / Watch] the following
[transcript / audio clip / video clip], taken
from a real conversation between two people.

These people are playing a game called
"Either Or," where they discuss whether they
would prefer one thing -- for example, the
ability to fly -- or an alternative, such as
the ability to breathe underwater.

Your job is to pay close attention to the
conversation and determine whether the two
people have met before -- meaning they are
familiar, such as friends, family members,
coworkers, or romantic partners -- or are
meeting for the first time in this
interaction -- meaning they are strangers.

For each clip, return:
  {
    "relationship_label": str {FAMILIAR|STRANGER},
    "confidence": float [0, 1],
    "reason": str
  }

Use the following confidence scale:
  0 = Not at all confident, ...,
  0.5 = Moderately confident, ...,
  1 = Extremely confident

Do not provide extra text outside the JSON
object.
\end{verbatim}
\end{small}

\section{Human Rating Study Details}
\label{app:study-details}

Raters were recruited on Prolific with platform-level quality screening in addition to the demographic criteria in Section~\ref{sec:human-ratings}: a Prolific approval rate of 99--100\% and at least 10 prior submissions. Under the planned-missing, 6-block design \citep{graham2006}, each rater was assigned to a single block and each of the 96 dyads appeared in two of the six blocks, so every dyad was rated by an overlapping subset of raters (25--35 per dyad). By design each rater judged 32 dyads; in the audio panel a platform issue left 13 of the 94 raters with 29--31 completed trials, while all text and video raters completed the full 32.

\paragraph{Inter-rater agreement.} Agreement among human raters is slight, as expected for a difficult perceptual task with a balanced label set. Fleiss's $\kappa$ (computed per stimulus over whichever raters saw it, accommodating the incomplete-block design) is $0.08$ for text and $0.17$ for both audio and video. Bennett's $S$ is essentially identical ($0.09$, $0.17$, $0.17$ for text, audio, video), as expected when the two classes are balanced \citep{gwet2021}. The near-zero text agreement is consistent with individual text raters performing at chance (Table~\ref{tab:comparison}): raters are not converging on a shared, reliable text cue.

\section{Rater Individual Differences}
\label{app:individual-differences}

Table~\ref{tab:tsis} reports, per modality, the reliability of the Social Information Processing (SP) subscale and its correlation with per-rater accuracy. The subscale is highly reliable in every panel, but its correlation with accuracy is null in audio and video and positive only in text, the modality where accuracy is at chance. We therefore do not interpret it as evidence of a general skill advantage.

\begin{table}[ht]
\centering
\small
\begin{tabular}{lcccc}
\toprule
Modality & $\alpha$ & Pearson $r$ & $p$ & Spearman $\rho$ \\
\midrule
Text  & 0.91 & $0.29$  & .004 & $0.23$ \\
Audio & 0.88 & $0.03$  & .78  & $0.00$ \\
Video & 0.92 & $-0.12$ & .24  & $-0.19$ \\
\bottomrule
\end{tabular}
\caption{SP subscale reliability and its correlation with per-rater accuracy ($N=94$ text, 94 audio, 92 video).}
\label{tab:tsis}
\end{table}

For self-reported cue use, raters most often reported attending to interactional cues (rapport, responsiveness) across all modalities, but individual cue-attendance scores rarely predicted accuracy (only a handful of per-cue correlations reached $p<.05$, with no stable cross-modality pattern). Full cue-use heatmaps are provided with the released analysis notebooks.

The per-rater accuracy clouds in Figure~\ref{fig:field} show the full distribution of per-rater accuracy in each modality (94 text, 94 audio, 92 video raters; each rater's accuracy over the 32 dyads in their assigned block, 29--32 for a few audio raters affected by a platform issue). The clouds are wide, but much of that width is sampling noise: with only $\sim$32 trials per rater, binomial variation around a common ability already reproduces most of the observed spread, including the upper tail that appears to exceed the best model (Section~\ref{sec:individual-differences}). The distribution should not be read as a stable ordering of raters by skill. This is also why per-rater accuracy is best modeled with partially-pooled crossed random effects (Section~\ref{sec:human-ratings}), which shrink noisy individual estimates, rather than trusting raw per-rater rates or treating each rater as a one-shot observation. Text raters cluster tightly around chance, consistent with the near-zero text agreement reported in Appendix~\ref{app:study-details}.

\section{Dyad Difficulty}
\label{app:dyad-difficulty}

Table~\ref{tab:difficulty} lists the eight hardest and eight easiest dyads by mean model-estimated probability of a correct human response, from the crossed random-effects model of Section~\ref{sec:dyad-difficulty}. Several of the hardest dyads fall below chance in every modality, functioning as systematic lures. Estimated easiness correlates across modality pairs (text--audio $r=0.52$, text--video $r=0.33$, audio--video $r=0.61$; all $p<.01$), indicating that difficulty is largely a property of the conversation rather than the observation channel.

Table~\ref{tab:humanmodeldiff} decomposes the human--model difficulty correlation reported in Section~\ref{sec:dyad-difficulty}. The raw pooled correlation (human-crowd accuracy vs.\ accuracy pooled over all models) is strong in audio, moderate in video, and only marginal in text. The weak raw text value is a suppression effect rather than absence of shared structure: humans and models carry opposing class biases (Section~\ref{sec:class-bias}), so their per-dyad accuracies partly anti-align on the familiar/stranger axis. Partialling out the true class raises the within-class correlation to $r=0.61$ (text), $0.58$ (audio), and $0.44$ (video), all $p<.001$. This within-class correlation measures fine-grained difficulty that is not reducible to a two-way class effect. The shared difficulty is, however, partly a crowd-level property: repeating the analysis with the single strongest model per modality in place of the pooled estimate leaves it robust only in video ($r=0.45$, $p<.001$), with weak raw associations in audio ($r=0.19$, $p=.06$) and text ($r=-0.04$, n.s.). Confidence-weighted single-model estimates match the binary ones almost exactly, so this is not an artifact of one prediction per dyad. Individual models thus express the human-aligned difficulty signal only noisily, and pooling recovers it.

\begin{table}[t]
\centering
\small
\begin{tabular}{lrrr}
\toprule
Estimator & Text & Audio & Video \\
\midrule
Pooled, raw & $0.20$ & $0.56$ & $0.34$ \\
Pooled, within-class & $0.61$ & $0.58$ & $0.44$ \\
Best model, raw & $-0.04$ & $0.19$ & $0.45$ \\
Best model, within-class & $0.24$ & $0.26$ & $0.47$ \\
\bottomrule
\end{tabular}
\caption{Human--model per-dyad difficulty correlation ($r$) under four estimators, by modality ($N=96$ dyads). ``Pooled'' averages model accuracy over all models; ``best model'' is the single most accurate model per modality (text \texttt{gpt\_text\_mini}, audio \texttt{gemini\_audio\_pro}, video \texttt{gemini\_video}). ``Within-class'' partials out the true familiar/stranger label.}
\label{tab:humanmodeldiff}
\end{table}

\begin{figure*}[t]
\centering
\includegraphics[width=\textwidth]{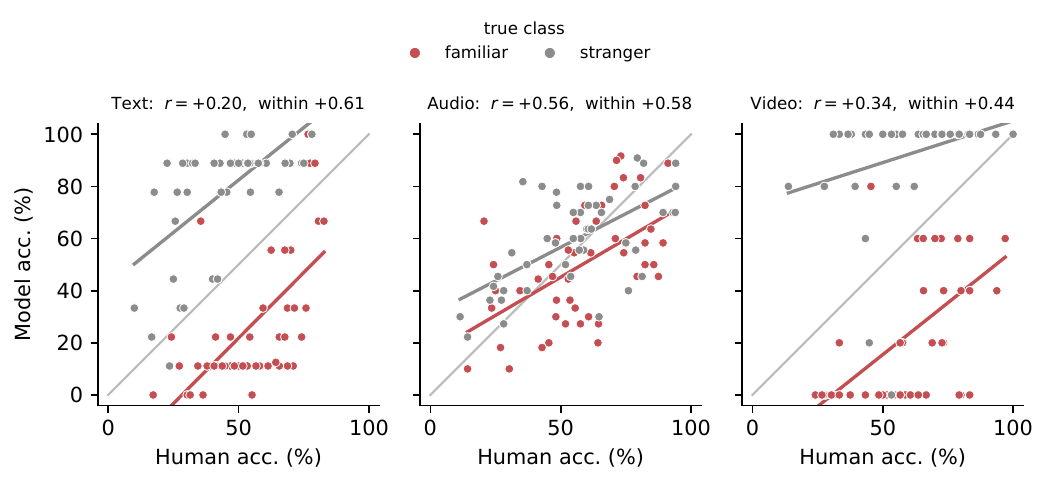}
\caption{Humans and models tend to find the same dyads hard (Section~\ref{sec:dyad-difficulty}). Each point is one of the 96 EO dyads, plotting human per-dyad accuracy (over all raters) against model per-dyad accuracy (pooled over all models), colored by true class; the diagonal marks equal difficulty. Per-dyad difficulty correlates positively in audio ($r=0.56$, $\approx31\%$ shared variance) and video ($r=0.34$, $\approx12\%$; both $p<.01$); text ($r=0.20$) is only marginally significant ($p=.045$).}
\label{fig:difficulty}
\end{figure*}

\begin{table}[ht]
\centering
\small
\begin{tabular}{llcccc}
\toprule
Dyad & Truth & Txt. & Aud. & Vid. & Mean \\
\midrule
\multicolumn{6}{l}{\emph{Hardest}} \\
\texttt{sample\_080} & str. & 29.6 & 32.0 & 22.2 & 27.9 \\
\texttt{sample\_055} & str. & 20.5 & 32.1 & 33.3 & 28.6 \\
\texttt{sample\_072} & fam. & 35.4 & 19.9 & 38.0 & 31.1 \\
\texttt{sample\_053} & str. & 31.1 & 27.0 & 40.7 & 32.9 \\
\texttt{sample\_038} & fam. & 25.8 & 28.5 & 47.9 & 34.1 \\
\texttt{sample\_057} & str. & 34.1 & 30.3 & 37.9 & 34.1 \\
\texttt{sample\_083} & str. & 31.7 & 34.7 & 36.1 & 34.2 \\
\texttt{sample\_065} & fam. & 34.9 & 33.4 & 35.0 & 34.4 \\
\midrule
\multicolumn{6}{l}{\emph{Easiest}} \\
\texttt{sample\_045} & fam. & 59.6 & 83.8 & 79.3 & 74.2 \\
\texttt{sample\_062} & str. & 69.2 & 75.2 & 79.4 & 74.6 \\
\texttt{sample\_051} & str. & 53.3 & 88.4 & 82.0 & 74.6 \\
\texttt{sample\_077} & fam. & 57.4 & 79.1 & 89.9 & 75.5 \\
\texttt{sample\_009} & fam. & 70.7 & 77.5 & 87.4 & 78.5 \\
\texttt{sample\_030} & str. & 63.3 & 83.7 & 91.1 & 79.4 \\
\texttt{sample\_073} & fam. & 71.4 & 86.1 & 81.9 & 79.8 \\
\texttt{sample\_016} & str. & 68.5 & 86.8 & 86.8 & 80.7 \\
\bottomrule
\end{tabular}
\caption{Hardest and easiest eight dyads by mean estimated P(correct human response) across modalities (\%). ``str.''\ = stranger, ``fam.''\ = familiar.}
\label{tab:difficulty}
\end{table}

\end{document}